\documentclass[authoryear,review]{elsarticle}
\usepackage[colorlinks=true,linkcolor=black, citecolor=blue, urlcolor=blue]{hyperref}
\usepackage{amssymb}
\usepackage{graphicx}
\usepackage{subfig}
\usepackage{svg}
\usepackage{doi}
\usepackage[font=footnotesize,labelfont=bf]{caption}
\usepackage[font=footnotesize,labelfont=bf]{subcaption}
\usepackage{lineno}
\usepackage[linesnumbered,ruled]{algorithm2e}
\usepackage{algpseudocode}
\usepackage[section]{placeins}
\usepackage[utf8]{inputenc}
\usepackage[T1]{fontenc}
\usepackage{xcolor}
\DeclareUnicodeCharacter{2212}{\textminus}
\DeclareUnicodeCharacter{2217}{*}
\usepackage{multirow}

\usepackage{lineno}

\journal{Computer and Electronics in Agriculture}

\usepackage[top=1in, right = 1in, left=1in]{geometry}

 \usepackage{float}
\begin{document}
\fontsize{12}{16}\selectfont
\begin{frontmatter}




\title{Machine Vision-Based Assessment of Fall Color Changes in Apple Leaves and its Relationship with Nitrogen Concentration }


\author[inst1]{Achyut Paudel\corref{cor1}}  
\ead{achyut.paudel@wsu.edu} 
\author[inst2]{Jostan Brown} 
\author[inst1]{Priyanka Upadhyaya} 
\author[inst1]{Atif Bilal Asad} 
\author[inst1]{Safal Kshetri}
\author[inst2]{Joseph R. Davidson}
\author[inst2]{Cindy Grimm}
\author[inst3]{Ashley Thompson}
\author[inst4]{Bernardita Sallato}
\author[inst5]{Matthew D. Whiting}
\author[inst1]{Manoj Karkee}

\affiliation[inst1]{organization={Biological Systems Engineering Department},
            addressline={Center for Precision and Automated Agricultural System, Washington State University}, 
            city={Prosser},
            postcode={99350}, 
            state={WA},
            country={USA}}
\affiliation[inst2]{organization={Collaborative Robotics and Intelligent Systems Institute},
            addressline={, Oregon State University}, 
            city={Corvallis},
            postcode={97331}, 
            state={OR},
            country={USA}}

\affiliation[inst3]{organization={Mid-Columbia Agricultural Research and Extension Center},
            addressline={, Oregon State University}, 
            city={Hood River},
            postcode={97031}, 
            state={OR},
            country={USA}}
\affiliation[inst4]{organization={Irrigated Agriculture Research and Extension Center},
            addressline={Washington State University}, 
            city={Prosser},
            postcode={99350}, 
            state={WA},
            country={USA}}
\affiliation[inst5]{organization={Department of Horticulture},
            addressline={Washington State University}, 
            city={Prosser},
            postcode={99350}, 
            state={WA},
            country={USA}}

\cortext[cor1]{Corresponding author.}


\begin{abstract}

Apple(\textit{Malus domestica} Borkh.) trees are deciduous, shedding leaves each year.  This process is preceded by a gradual change in leaf color from green to yellow as chlorophyll is degraded prior to abscission. The initiation and rate of this color change are affected by many factors including leaf nitrogen (N) concentration.  We predict that leaf color during this transition may be indicative of the nitrogen status of apple trees. This study assesses a machine vision-based system for quantifying the change in leaf color and its correlation with leaf nitrogen content. An image dataset was collected in color and 3D   over five weeks in the fall of 2021 and 2023 at a commercial orchard using a ground vehicle-based stereovision sensor. Trees in the foreground were segmented from the point cloud using color and depth thresholding methods. Then, to estimate the proportion of yellow leaves per canopy, the color information of the segmented canopy area was quantified using a custom-defined metric, “\textit{yellowness index}” (a normalized ratio of yellow to green foliage in the tree) that varied from -1 to +1 (-1 being completely green and +1 being completely yellow). Both K-means-based methods and gradient boosting methods were used to estimate the \textit{yellowness index}. The gradient boosting based method proposed in this study was better than the K-means-based method (both in terms of computational time and accuracy), achieving an $R^2$ of 0.72 in estimating the \textit{yellowness index}. The metric was able to capture the gradual color transition from green to yellow over the study duration. Trees with lower leaf nitrogen showed the color transition to yellow earlier than the trees with higher nitrogen. The onset of color transition during both years occurred during the $29^{th}$ week post-full bloom (October 22 in 2021 and Nov 10 in 2023). This critical timing could be used for conducting nitrogen status analysis on apple trees using machine vision, enabling more precise and timely assessment of nutrient levels and facilitating targeted fertilization strategies in orchard management. 

\end{abstract}
\begin{keyword}
Leaf Color Change \sep Machine Vision \sep  Point Cloud Segmentation \sep Precision Nitrogen Management \sep Machine Learning
\end{keyword}

\end{frontmatter}


\section{Introduction}

Water and nutrient applications in commercial tree fruit orchards are managed at a block or field level, providing treatment relatively uniformly to hundreds or even thousands of trees  without considering the variability among individual trees. This homogeneous management of nutrients in tree fruit crops often leads to sub- and/or super-optimal inputs, with negative impacts on tree growth, fruit yield, and fruit quality \citep{klein1989drip, neilsen2009nitrogen}. Managing these inputs at the individual tree level (i.e. addressing the unique needs of individual trees)  is a promising concept for improving tree vigor, fruit yield and quality. However, managing orchards at the individual tree level is challenging, particularly in terms of assessing the needs of individual trees and making precise management decisions. Each tree within an orchard is unique with its specific needs for water, nutrients, and other inputs \citep{aggelopoulou2010spatial, james_a_taylor_spatial_2007}. For eg., \citet{aggelopoulou2010spatial} reported spatial variability with a coefficient of variation (CV) of up to 82\% for bloom and 40\% for yield within the same block. The variability among trees can be caused by multiple factors including spatial variation of soil properties like texture, depth, slope, or microbial activities, variation in microclimates within orchards, and differences in tree training and pruning \citep{aggelopoulou2010spatial}. Effective nutrient management at the individual tree level could be possible if these multiple factors can be accurately assessed for each individual tree.

Nitrogen (N) is a major macro nutrient needed by apple (\textit{Malus domestica} Borkh.) trees given its fundamental roles in tree growth, canopy and crop development, and fruit quality \citep{sanchez1995nitrogen,neilsen2002efficient}. Traditionally, the most common approach farmers use to assess tree N status involves laboratory-based assessments of leaf tissue.  These can provide a quantitative indicator of uptake, and when contrasted with a standard provide an estimation of sufficiency \citep{marschner2011marschner}. Soil chemical testing of nitrate (NO$_3^-$) or ammonium (NH$_4^+$) availability, on the other hand, is not a reliable method to determine N supply, given its variability across the season and multiple forms of losses \citep{fox1978field}. These destructive methods demand significant time for sampling and their high cost limits the sampling density to few trees or locations, a practice that lacks precision due to inherent spatial variations within orchards \citep{arno_comparing_2017}.

In addition to laboratory-based tests, experienced farmers and researchers often employ visual assessment of canopy features such as shoot length, canopy density, trunk diameter, overall tree growth, crop load, and leaf color to estimate N status.. While these observational methods leverage professional expertise, they suffer from subjectivity, and potential variability among assessors. The reliance on seasoned professionals further limits its widespread applicability in commercial orchards.

These limitations of both laboratory-based methods and visual assessments by subject matter experts underscore the need for a rapid, reliable, and more comprehensive approach to N status evaluation \citep{cheng_nitrogen_2004} at the individual tree level. A promising alternative is a machine-vision system that automates and standardizes the quantification of visual parameters across entire orchards. This approach could provide objective, empirical data that may be correlated with tree nitrogen status through modeling methods. By enabling more precise, tree-level management in commercial orchards, such a system has the potential to overcome the sampling and subjectivity issues inherent in current methods while offering scalability for broader applications.

Machine vision techniques have been extensively used to evaluate various canopy-level features and detect canopy stresses in fruit crops \citep{wang2023automatic,paudel2022canopy,wang2018machine,brown2024tree}. More specifically, many recent studies have reported machine vision-based tools for estimating canopy density\citep{mahmud2021development, paudel2023vision}, tree trunk cross-sectional area \citep{wang2023automatic}, and canopy color \citep{naschitz2014effects} in fruit trees.

Leaf color is one of the key canopy characteristics that is correlated with nitrogen content\citep{ali2012leaf, ye2020estimation}. This correlation is due to the presence of four rings of N in the chlorophyll, the pigment in leaves responsible for light harvesting\citep{marschner2011marschner}. Chlorophyll reflects light primarily in the green spectrum, which is why plant tissue containing chlorophyll appears green. In deciduous trees like apples, a significant color change occurs from green to yellow as the growing season progresses towards fall, as the chlorophyll breaks down and mobile elements like N move back into the perennial structures of tree. This relocation of nitrogen from leaves to woody regions for storage will later support next seasons’ growth\citep{murneek1932autumnal,neilsen2002efficient}.A greener canopy with vigorous growth is often correlated with a tree having high nitrogen, while a yellower and less vigorous canopy is often associated with trees with low nitrogen \citep{raese2007nitrogen}.  In addition, trees with high nitrogen tend to maintain their green color longer into the season providing a potential indication of tree nitrogen status during this critical transition.  

Various RGB, multispectral, and hyperspectral imaging techniques have been evaluated to quantify the color and estimate nitrogen or chlorophyll content in apples \citep{neilsen1995using,perry2007spectral}. Most of these studies use devices like chlorophyllmeter (SPAD) or satellite images.   While handheld devices like SPAD meters can be utilized in the field, they pose challenges for scaling assessments across entire orchards due to the need for data collection from individual leaves. On the other hand, satellite imagery offers broader insights but often lacks the resolution required to pinpoint individual trees, necessitating further computations for accurate matching. Our study overcomes these limitations by focusing on individual tree-level assessments, providing a scalable and efficient approach for evaluating apple tree health and productivity in orchards.

In this study, we investigated the relationship between N status and color degradation utilizing RGB-D camera during apple tree senescence, . and explored the potential to utilize this assessment for precision management practices at the individual tree level. 

Specifically, the study had the following two objectives:
\begin{enumerate}
    \item Develop a machine vision-based technique to quantify foliage color during late fall in an apple orchard.
    \item Use the foliage color as an indicator to differentiate between trees with varying foliage leaf nitrogen concentration.
\end{enumerate}


\section{Materials and Methods}\label{sec:materials_and_methods}
\begin{figure}[!h]
    \centering
    \includegraphics[width=0.4\textwidth]{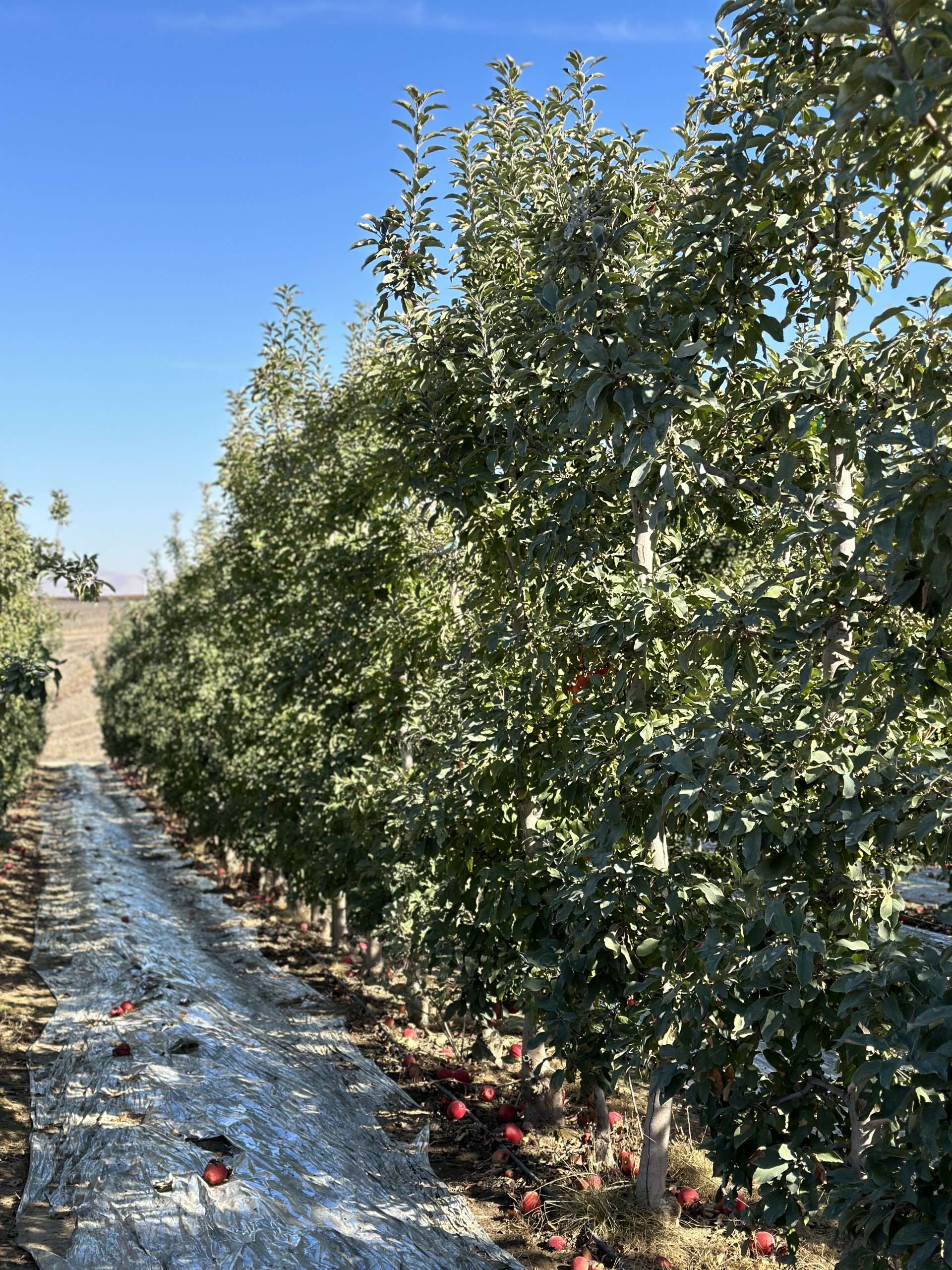}
    \caption{ A representative high density apple orchard plantation in Prosser, Washington. The trees have been trained in a tall spindle architecture and spaced closely (~4 foot) forming a highly dense planar “wall”.}
    \label{fig:high_density}
\end{figure}

\begin{figure}[!h]
\centering
{\includegraphics[width=15cm]{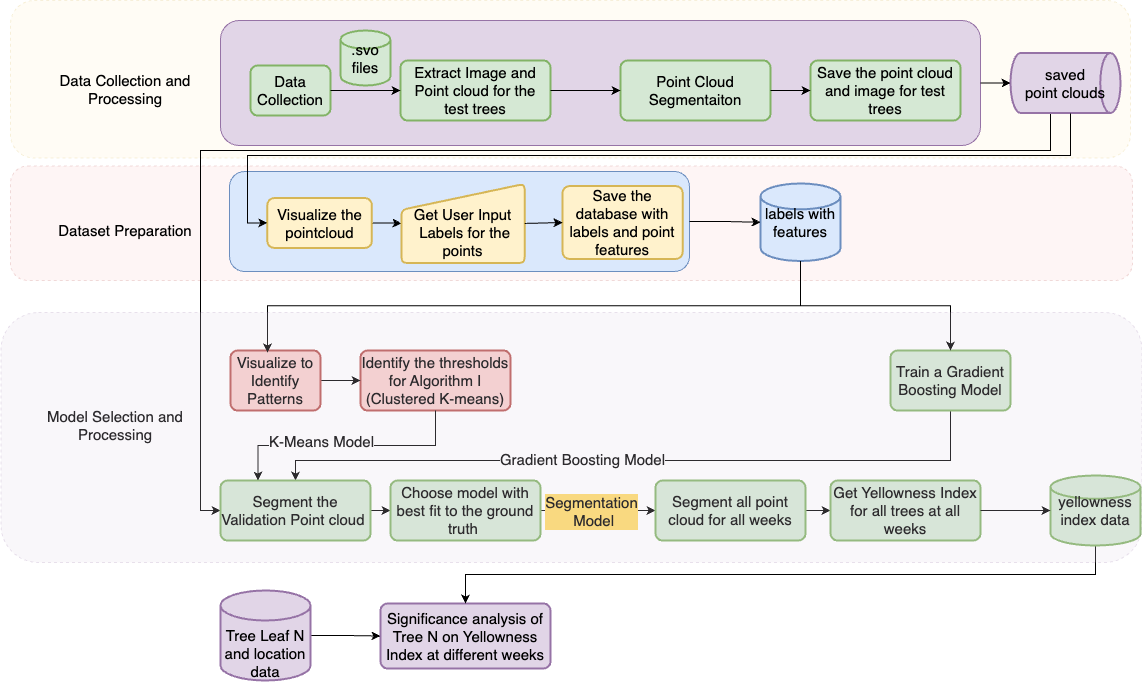}}
\caption{Overall data collection and processing steps used in the study; Each light-colored box represents a distinct process and has been described further in the subsequent sections.}
\label{fig:methods}
\end{figure}

RGB-D images were collected over several weeks in fall of 2021 and 2023. These images were collected in a Jazz\textsuperscript{\texttrademark} commercial orchard (Yakima Valley Orchard) near Prosser, WA, USA.Data was collected from a commercial apple orchard located in Prosser, WA (Figure \ref{fig:high_density}), where jazz apple trees were grafted onto M9 rootstock. The trees were planted in 2008 and trained using a tall spindle, vertical fruiting wall system. They were spaced closely at 1.4 meters (4.5 feet) apart, with 2.6 meters (8.5 feet) between adjacent rows. The irrigation system included a combination of drip lines and both overhead and under-canopy irrigation, laid out uniformly throughout. The soil composition consists of silt loam over a layer of calcium carbonate, approximately 1-2 inches thick, which sits atop basalt. The point clouds of the tree canopies were then segmented into yellow and green foliage; a quantitative metric, termed the \textit{yellowness index}, was created to estimate the color of the trees. The yellowness indices of the trees during different weeks were then correlated with the leaf nitrogen content for each year. The overall data collection and processing method used in this study is shown in Figure \ref{fig:methods}.

\subsection{Data Collection}

The datasets (color and 3D images) for this study were collected over five weeks in fall, 2021 (from October 13 to November 14) and fall, 2023 (from October 20 to November 17), every week. The time to start the collection of data was determined by when the leaves began to change color from green to yellow (Figure \ref{fig:trend_images_23}) upon consultation with the orchard manager. The timing of this is dependent on a variety of factor including temperature throughout the growing season, the location of the orchard, and other management practices in the orchard. The dataset was collected from a total of 200 trees across 17 rows of the test orchard, with every other 3rd tree in each row taken as a sample. The data were collected using a commercial stereovision-based RGB-D camera (Zed2i, Stereolabs, Paris, France) mounted on a utility vehicle (Figure \ref{fig:gator,location,orchard}b). The camera was mounted ~2 meters above the ground to ensure that most of the canopy (Figure \ref{fig:trend_images_23}, part of tree above ~0.3m from ground) was visible. An SVO file (Stereolabs file format) with all the metadata including the camera parameters, images, and sensor data was recorded for each row of trees. Images (pixel resolution of 2208 x 1242) and point clouds (maximum depth of 10m) for each tree were later extracted based on timestamp matching between frames and user input. During processing, the RGB and co-registered point cloud were identified for each test tree.

\begin{figure}[!h]
\centering
{\includegraphics[width=0.8\textwidth]{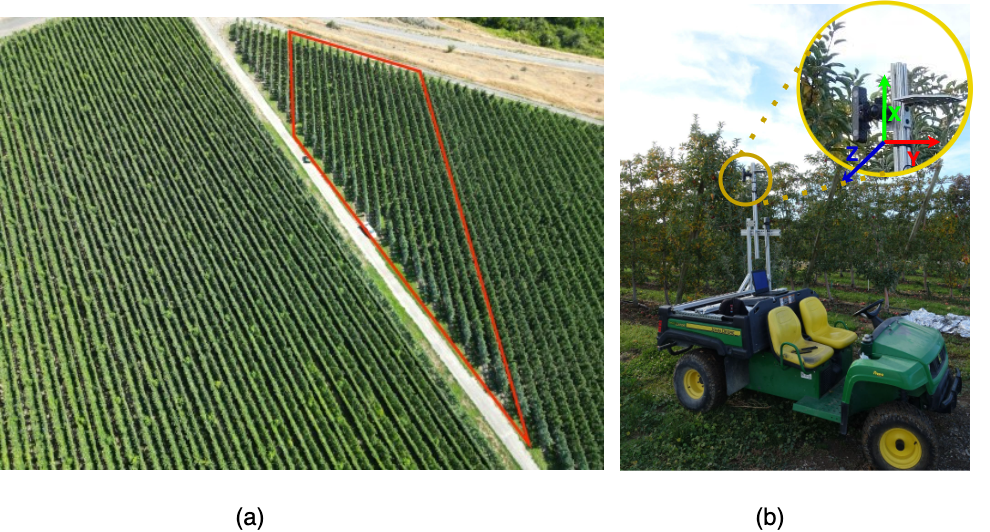}}
\caption{Study site and sensing setup; (a) An aerial view of the test plot. The red outline shows the location of the plot used in this study., and b) Ground vehicle with a camera mounted on top (zoomed portion shows camera and axes orientations)}
\label{fig:gator,location,orchard}
\end{figure}

\begin{figure}[!h]
\centering
{\includegraphics[width=15cm]{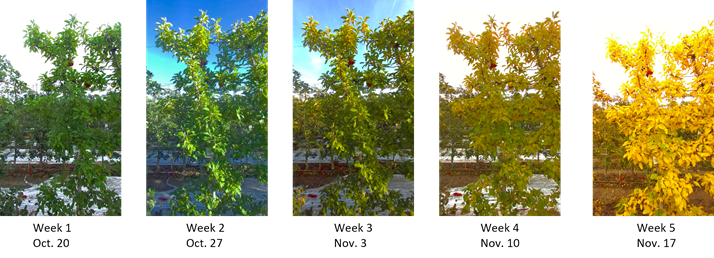}}
\caption{Images of a sample apple tree acquired over the data collection period in 2023. The foliage can be seen gradually changing from green to yellow.}
\label{fig:trend_images_23}
\end{figure}

\subsection{Point Cloud Processing and Segmentation}

\subsubsection{Foreground Tree Segmentation}\label{sec:tree-seg}

The initial data acquired through the stereovision sensor were noisy and required further refinement. An initial step involved applying a color threshold of 153 to the blue channel to filter out pixels corresponding to the sky in the RGB color space. Subsequently, a distance threshold of 3 meters along the Z-axis (depth axis), was implemented. The optimal threshold values were determined experimentally through trials and visual inspections. Consequently, the resulting point cloud consisted of points mostly from the foreground tree, with some residual points from the ground. To eliminate the undesirable ground points, the points within 50 centimeters above the lowest x-value (along the height of the tree) were excluded. The resulting point cloud was down-sampled by selecting every tenth point, effectively reducing its size while preserving essential information for subsequent analysis.

\subsubsection{Color Segmentation and Clustering} \label{sec:clustering}

\begin{figure}[!h]
    \centering
    \includegraphics[width=\textwidth]{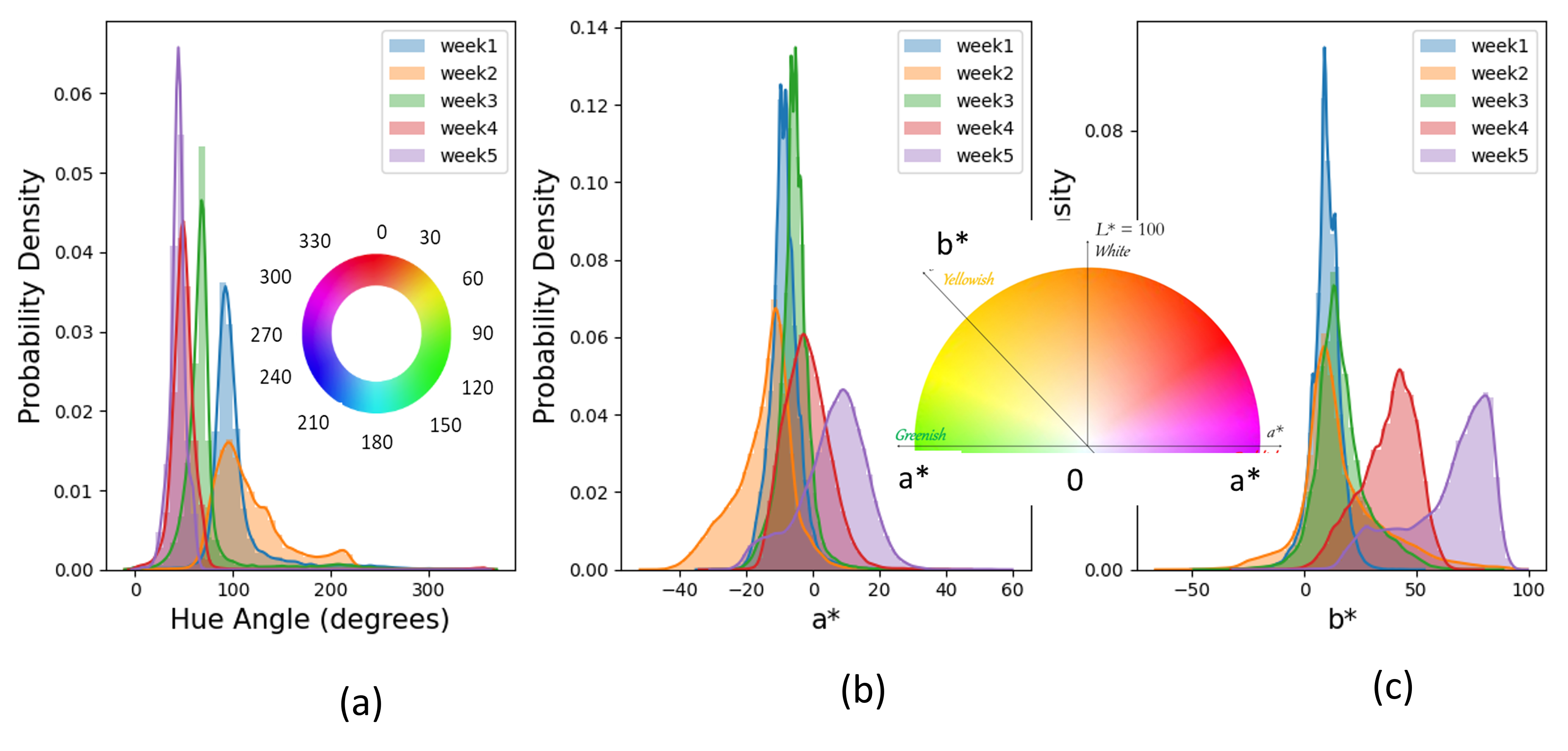}
    \caption{Distribution of a) Hue angles and b) a* and c) b* values for one of the sample trees during the five-week study period of 2023. The color chart shows the color associated with different values for (a) hue angles, (b) \citep{thompson_how_2017} a*, and b*. The hue angle decreased in a) and a* and b* increased in b) and c) as the weeks progressed during the study all signifying the shift from green to yellow.}
    \label{fig:hue_ab_shift_2023_569}
\end{figure}

The point cloud obtained from step \ref{sec:tree-seg} consisted of points from the foreground tree, and could be broadly categorized into yellow leaves, green leaves, and the trunk.  To quantify the color changes in trees’ over the duration of study, HSV (Figure \ref{fig:hue_ab_shift_2023_569}a) and CIE-Lab* (Figure \ref{fig:hue_ab_shift_2023_569}b and \ref{fig:hue_ab_shift_2023_569}c) color spaces were used as they are less affected by the variation in intensities. In the HSV space, hue remains unaffected by varying light intensity, while in the CIE-Lab* space, the a* and b* values correspond respectively to red-green and blue-yellow colors. RGB images captured by the RGB-D sensor were converted to both the CIE-La*b* (Lab) space, using a standard D65 illuminant, and the HSV color space. These conversions were performed using the scikit-image library \citep{van2014scikit}.

The transition of the trees from green to yellow color was noticeable in both HSV and La*b* space (Figures \ref{fig:hue_ab_shift_2023_569}a and \ref{fig:hue_ab_shift_2023_569}b,c). A probability density plot for the hue space across the weeks illustrated a decline from approximately 110 degrees in week 1 to about 50 degrees in week 6 (hue expressed as degrees between 0 and 360, as shown in Figure \ref{fig:hue_ab_shift_2023_569}a), signifying the shift from green to yellow. A similar trend was visible in the probability density plots for a* and b* in the La*b* space. Both a* and b* showed a slight increase as the weeks progressed, corresponding to a color change from green to yellow.

The transition in tree canopy color was apparent qualitatively, but a robust quantitative approach was needed to precisely quantify them. To achieve this, individual points in the point cloud were categorized into distinct groups of yellow leaves, green leaves, or the trunk. Two different approaches were used: a) An unsupervised K-means based method and b) A supervised Gradient Boosting Method.

\textbf{\textit{K-Means Based Segmentation}}: Qualitative analyses using K-means clustering over multiple samples in both the La*b* and HSV color spaces showed that the La*b* space performed better in distinguishing yellow and green pixels as separate classes. Consequently, the subsequent analysis was conducted using La*b* color space. To cluster the point cloud, a modified k-means algorithm (Algorithm 1) was applied, initially grouping the points into 20 clusters. A threshold for the cluster center was set in both the a* and b* space, to merge these clusters into three distinct classes: $Yellow$, $Green$, and $Trunk$. The $Yellow$ cluster encompassed foliage that had transitioned to yellow, while the $Green$ cluster comprised foliage that retained its green color. The $Trunk$ cluster included points from the trunk, branches, background soil, and some brown leaves. A detailed breakdown of the clustering process is outlined in Algorithm \ref{alg:clustering}.

\SetAlgoNoLine
\begin{algorithm}[H] 
    \DontPrintSemicolon
    \KwIn{Point cloud from stereovision system, $P$}
    \KwOut{Clustered Point cloud--3 groups: Green ($c_g$), Yellow ($c_y$), Trunk ($c_t$)}

    Define $n$ as the initial number of clusters (integer)\;
    Perform K-Means clustering using $a*$, $b*$ values into $n$ clusters ($C = \{c_1, c_2, ..., c_n\}$)\;
    
    Initialize Green cluster ($c_g$), Yellow cluster ($c_y$), and Trunk cluster ($c_t$) as empty sets: \;
    $c_g = \emptyset$; $c_y = \emptyset$; $c_t = \emptyset$\;
    
    Define $a*$ and $b*$ thresholds for Green cluster: $[cg_{amin}, cg_{amax}]$, $[cg_{bmin}, cg_{bmax}]$\;
    Define $a*$ and $b*$  thresholds for Yellow cluster: $[cy_{amin}, cy_{amax}]$, $[cy_{bmin}, cy_{bmax}]$\;
    Define $a*$ and $b*$  thresholds for Trunk cluster: $[ct_{amin}, ct_{amax}]$, $[ct_{bmin}, ct_{bmax}]$\;

    \For{$i \in \{1, 2, ..., n\}$}{
        \uIf{$cg_{amin} \leq c_i(a) \leq cg_{amax}$ \textbf{and} $cg_{bmin} \leq c_i(b) \leq cg_{bmax}$}{
            $c_g = c_g \cup c_i$\; \tcp{Merge cluster $c_i$ into Green cluster}
        }
        \uElseIf{$cy_{amin} \leq c_i(a) \leq cy_{amax}$ \textbf{and} $cy_{bmin} \leq c_i(b) \leq cy_{bmax}$}{
            $c_y = c_y \cup c_i$\; \tcp{Merge cluster $c_i$ into Yellow cluster}
        }
        \ElseIf{$ct_{amin} \leq c_i(a) \leq ct_{amax}$ \textbf{and} $ct_{bmin} \leq c_i(b) \leq ct_{bmax}$}{
            $c_t = c_t \cup c_i$\; \tcp{Merge cluster $c_i$ into Trunk cluster}
        }
    }
    \Return $c_g$, $c_y$, $c_t$
    \caption{Clustering of Point Cloud based on $a*$, $b*$ values}
    \label{alg:clustering}
\end{algorithm}

\begin{figure}[!h]
    \centering
    \includegraphics[width=0.6\textwidth]{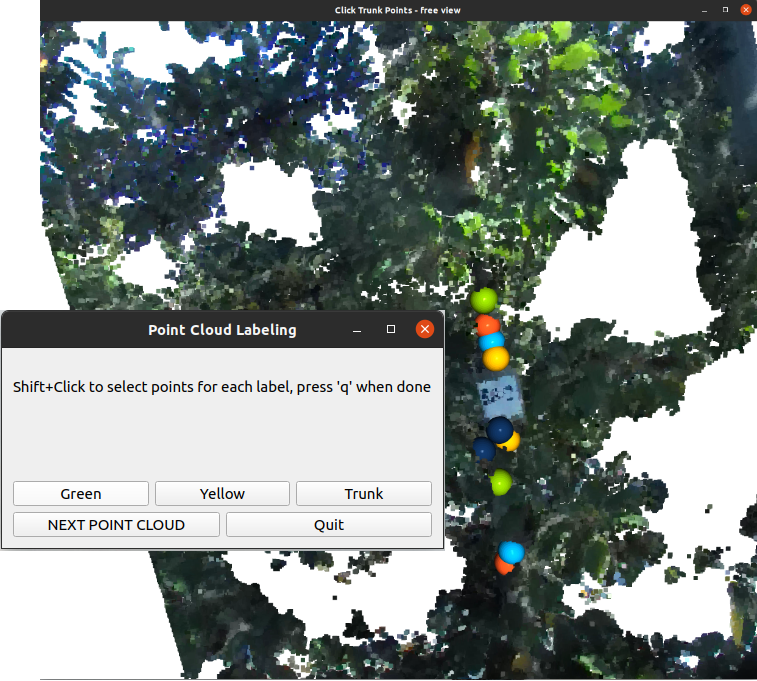}
    \caption{A custom program to select the points belonging to multiple classes - Green, Yellow, and Trunk. This instance showed an interactive window where the user selected points belonging to the trunk. The features of the points - including a*,b*,R, G, B, and their eigenvector and eigenvalues (in a neighborhood) were recorded into a spreadsheet for further analysis.}
    \label{fig:valid_window}
\end{figure}

In the aforementioned algorithm, the manual setting of threshold values for a* and b* required multiple iterations to identify optimal values suitable across the data throughout the weeks. To standardize this process, a methodology involving manual user input labels was implemented to obtain a labeled dataset. A custom Python program (Figure \ref{fig:valid_window}), using the Open3D library \citep{zhou2018open3d}, was developed to label the point cloud. This interactive program allowed users to choose between three labels - ’Green’, ’Yellow’, and ’Trunk’ and manually select the points belonging to these labels. Trees were randomly sampled throughout the entire data collection period, and the labels along with the characteristics of the points were recorded to create the dataset.

\begin{figure}[!h]
    \centering
    \includegraphics[width=12cm]{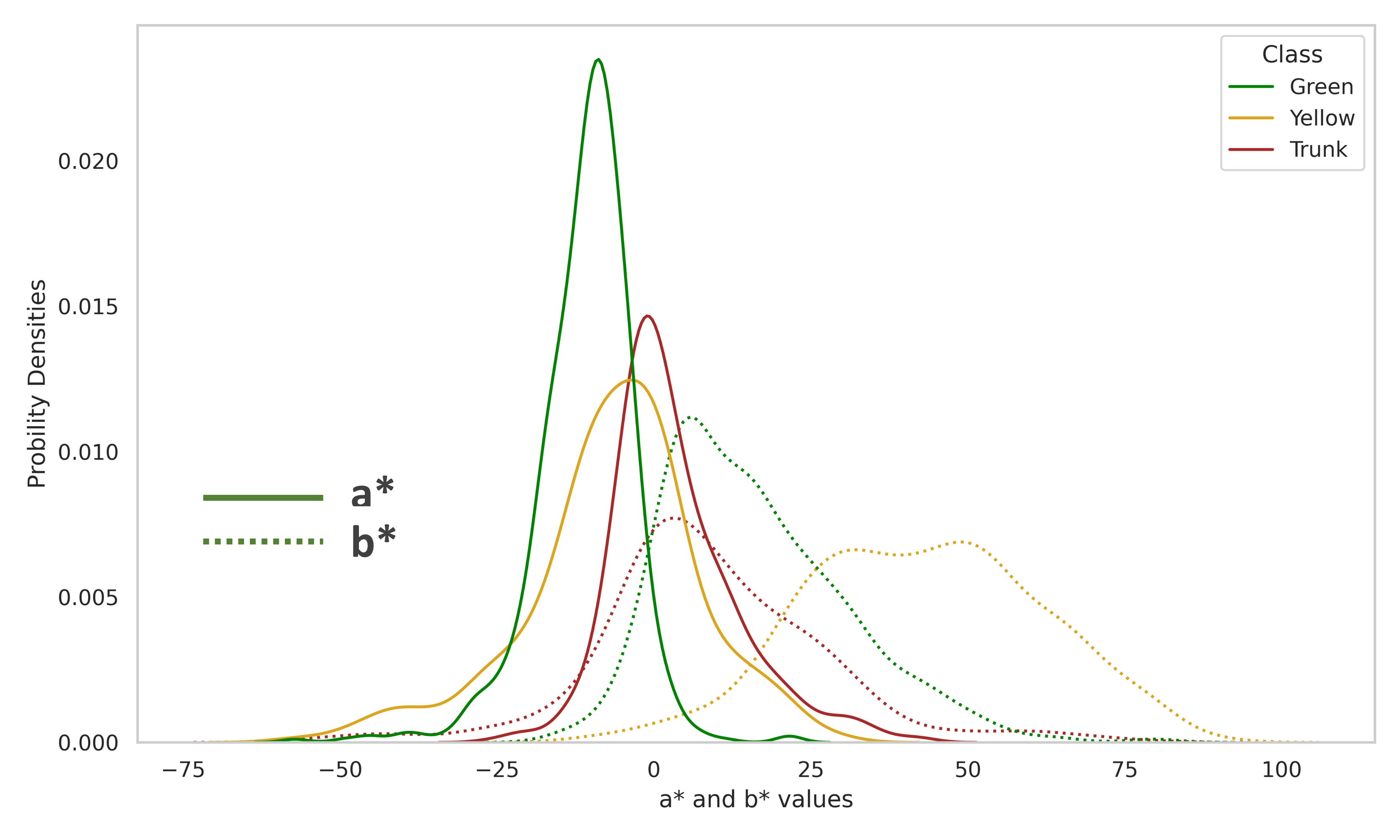}
    \caption{a* and b* values in CIE-L*a*b* space during different weeks of image acquisition (2023) for points belonging to green, yellow, and trunk classes. The solid lines represent a* values, and dotted lines represent b* values, different colors represent different classes.}
    \label{fig:ab_pattern_2023}
\end{figure}

To visualize the difference between groups, a distribution plot of the a*, and b* spaces was created (Figure \ref{fig:ab_pattern_2023}). These plots provided strategic reference points for selecting threshold values. For the 2023 dataset, the threshold values for a* and b* were determined as follows: $a^* < -10$, $ 0 < b^* < 25$ for $Green$, $b∗ > 45 $for $Yellow$, and $a∗ > 0, 0 < b∗ < 50$ for $Trunk$. While K-means clustering, being an unsupervised method, doesn’t necessitate user inputs for training data and attempts to identify inherent clusters in the dataset, it comes with certain drawbacks. The method’s main limitations include longer computation times due to the need to process all points and its reliance on random initialization, potentially leading to different solutions with each run.

\textbf{\textit{Gradient Boost Classifier Based Segmentation}}: An alternative approach utilized a gradient boost classifier, an ensemble learning technique that combines multiple weak learners (typically decision trees) to create a stronger, more robust model \citep{bartlett1998boosting}. It sequentially builds new models, focusing on correcting the errors made by the previous ones, ultimately producing a more accurate and powerful predictive model. In this study, the gradient boost classifier model was created using scikit-learn \citep{van2014scikit}. The dataset was divided into 80\% training and 20\% testing sub-sets. The model required tuning of different hyperparameters. A strategic approach to varying model hyperparameters was employed, with one hyperparameter held constant while varying the others to analyze their individual effects on model accuracy. The hyperparameter learning rate was varied between 0.1-1, max depth was varied between 1-5, and n estimator was varied between 100-1000 and their performance on training and test dataset was analyzed. The set of hyperparameters that achieved the highest accuracy on the test data was then selected.

The outputs from both Algorithm 1 (K-Means) and the Gradient Boosting method provided labels—’Green’, ’Yellow’, or ’Trunk’—for each point in the point cloud. Following the clustering of points, a metric, \textit{yellowness index}, was used to quantify the extent of yellowing within each tree canopy. The \textit{yellowness index}, calculated using Equation \ref{eq:yel}, represents the normalized ratio between yellow and green points within a tree canopy, with values ranging between -1 and +1. A value of -1 indicates that the tree was entirely green, while a value of +1 signifies complete yellowing.

\begin{equation} \label{eq:yel}
\begin{array}{ll}
    \mbox{\textit{\textit{yellowness index}}}  = \frac{{y - g}}{{y + g}}
\end{array}
\end{equation}

 \par \noindent where,\\
 $y$ = number of points with 'Yellow' label ($c_y$)\\
 $g$ = number of points in 'Green' label ($c_g$)\\

To assess the performance of the yellowness estimation techniques, 50 validation trees were randomly selected and deleafed between the second and fourth trellis wires (a section of the trees, Figure 9). The leaves removed from each tree were collected in a bag and then later segregated manually into separate groups (“Green” or “Yellow”) for each tree. A leaf was identified as ’Green’ if more than half of the leaf area was green, and ’Yellow’ if otherwise. The precise weights of the yellow and green leaf groups were measured using a highly accurate Mettler PC 4000 (Mettler Toledo, Switzerland) weighing scale. This value served as the ground truth to assess the models’ performance. The \textit{yellowness index} value was computed for all validation trees using Equation \ref{eq:yel}. Both Algorithm 1 (K-means) and Gradient Boosting were evaluated against the ground truth data, with the superior-performing model based on this evaluation selected for further study.  A timing test was also performed to evaluate the computation time of each of the model. Both models were run on a System76 Oryx Pro (System76, Inc., Colorado, USA) equipped with 16GB RAM, an 8GB NVIDIA GeForce RTX 3070 GPU (NVIDIA Corp., California, USA), and an Intel i7-11800H processor (Intel Corp., California, USA) at 2.30 GHz with 16 cores.

\begin{figure}[!h]
    \centering
    \includegraphics[width=8cm]{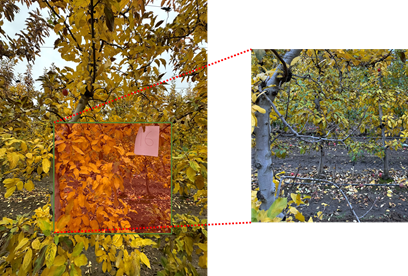}
    \caption{Images of a sample tree before and after deleafing. The leaves of 50 sample trees within a section between the second and fourth trellis wire were removed (shown in transparent red box) and collected to estimate the ground truth \textit{yellowness index}.}
    \label{fig:valid_image_23}
\end{figure}

\subsection{Yellowness Index and Leaf Nitrogen Concentration}\label{sec:nitrogen_relation_method}

To investigate if the \textit{yellowness index} had any relationship with the foliar nitrogen concentration on the trees, the \textit{yellowness index} values were regressed against the leaf nitrogen concentration. The leaf nitrogen concentration was obtained using the standard Kjeldahl’s method\citep{kirk1950kjeldahl} by collecting 50 fully grown mid-shoot leaves from each tree, at the end of the growing season (Late July in 2021 and early August in 2023). The leaf nitrogen concentration represents the average leaf nitrogen concentration of each tree and is associated with canopy growth and canopy color. The leaf nitrogen level of 2-2.4\% was regarded as an adequate range for nitrogen concentration based on referenced values for apples\citep{cheng_nitrogen_2004, bernardita_nitrogen_2021}.

A one-way ANOVA was performed to examine differences in the yellowness index among trees. This analysis was conducted separately for each week of the season, comparing trees grouped by their leaf nitrogen levels to determine if trees with different nitrogen statuses exhibited significant variations in their yellowness index throughout senescence. The trees were classified into 5 groups according to their leaf nitrogen concentrations: 'VeryLow'($N<1.7\%$), 'Low'($1.7\%< N<2\%$), 'Good'($2\%< N<2.4\%$), 'High'($2.4\%<N<2.6\%$), and 'VeryHigh'($ N >2.6\%$). The ANOVA analysis was followed up by a Tukey HSD posthoc analysis to analyze the significant difference between pairs of groups if ANOVA showed a significant difference between those groups.


\section{Results and Discussions}\label{sec:results}

\subsection{Tree Segmentation and Clustering}\label{sec:tree_seg_result}

\begin{figure}[!h]
    \centering
    \includegraphics[width=\textwidth]{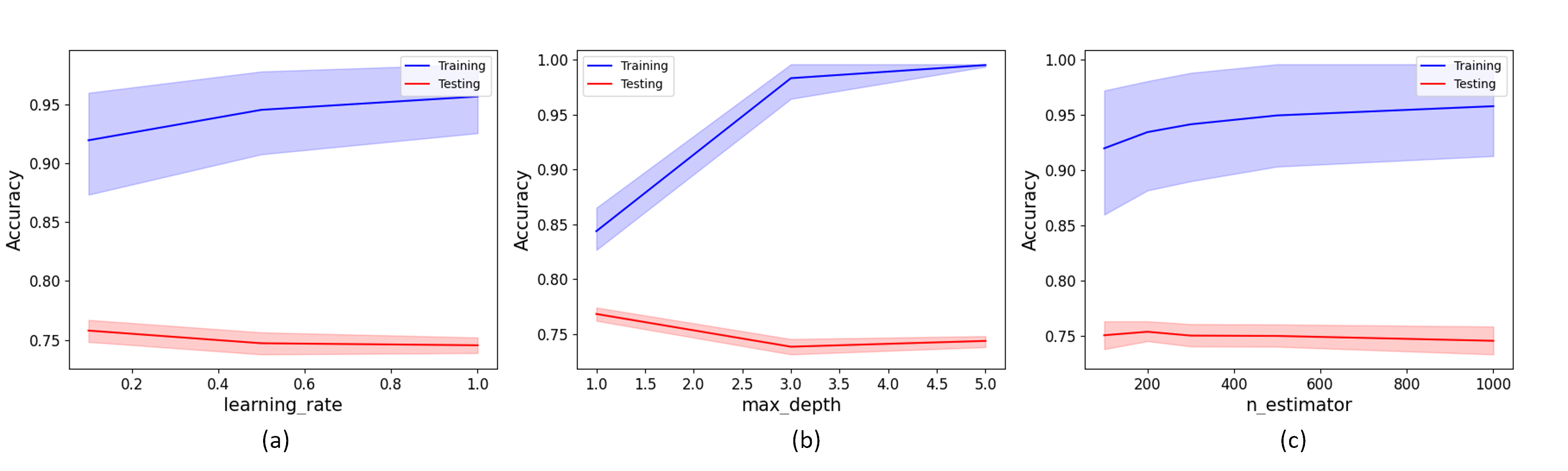}
    \caption{Varying accuracies for the gradient boost model with different combination of hyperparameters (a) Learning Rate, b) Maximum Depth of Tree, and c) Number of Estimators. The blue line denotes varying accuracy during training whereas the red line represents varying accuracy during testing. The shaded regions around the lines illustrate the range of accuracies across different values of the other hyperparameters, with the indicated hyperparameter being held constant.}
    \label{fig:hyperparameter}
\end{figure}

The original color point cloud was clustered into green and yellow groups to quantify the yellowness of each tree canopy in the test orchard. As discussed in the methods section, Algorithm 1 (K-means based) and the Gradient Boost classifier were used to classify leaves into these two groups. For the Gradient Boost model, \textit{learning rate}, \textit{max depth}, and \textit{n estimator} were varied to identify the best set of hyperparameters (Figure 11). The model started to overfit the training data on increasing the hyperparameters, as indicated by higher performance on the training set with reduced performance on the test set (e.g., training accuracy went from 0.85 to nearly 1, while testing accuracy decreased as \textit{$max\_depth$} of the trees was varied from 1 to 3, Fig \ref{fig:hyperparameter}b). The gradient boost method with the best set of hyperparameters achieved an accuracy of 78\% on the test set with a \textit{learning rate}e of 0.1, \textit{maximum depth} of 1, and \textit{number of estimators} as 100.

\begin{figure}[!h]
    \centering
    \includegraphics[width=10cm]{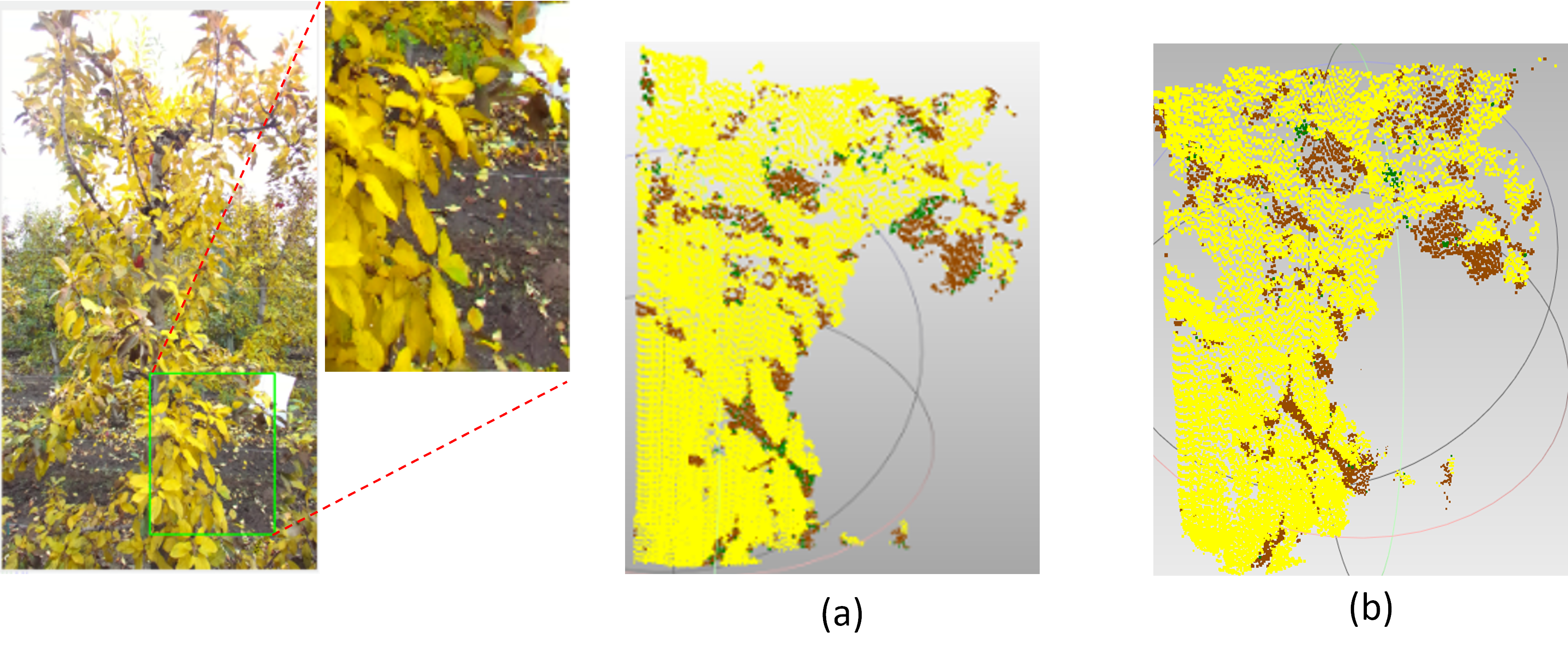}
    \caption{Example segmented point cloud of tree canopies in the validation data. The region delineated by the green rectangle was cropped and the corresponding color image and point cloud data was extracted for performance assessment. The results from the segmentation models on the cropped point cloud are shown in a) Algorithm 1; and b) Gradient Boost Classifier. The three colors in the point cloud represent three groups; green leaves, yellow leaves, and trunk.}
    \label{fig:yellowness_valid}
\end{figure}

\begin{figure}[!h]
    \centering
    \includegraphics[width=\textwidth]{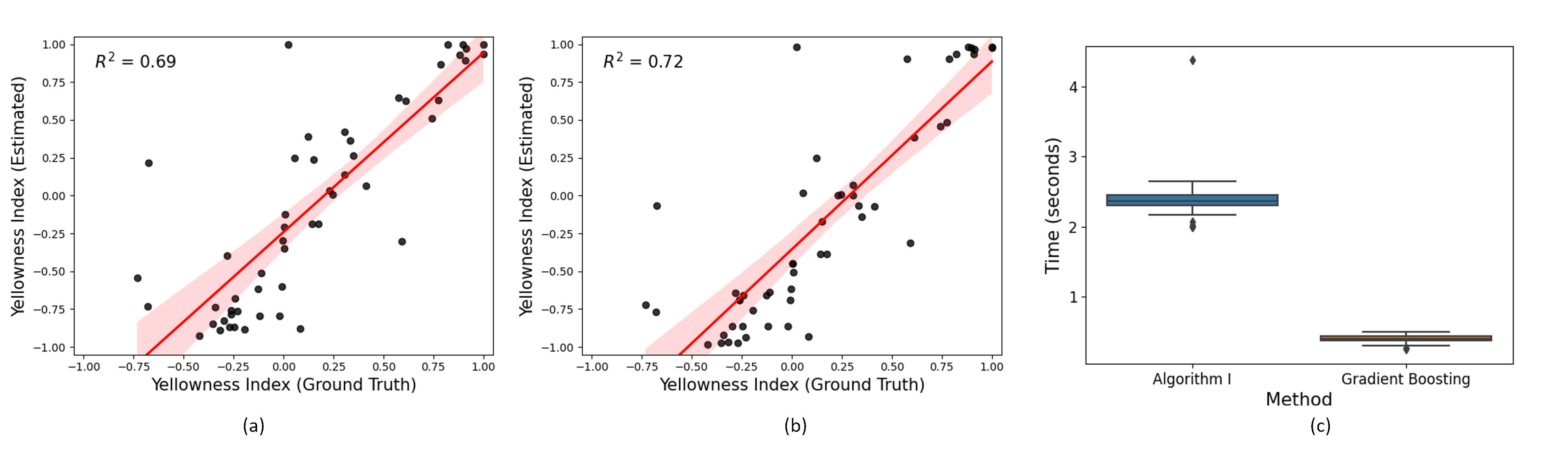}
    \caption{Relationship between estimated and manually calculated yellowness index for two classification/clustering models on the validation dataset; estimated vs ground truth \textit{yellowness index} achieved by a) Algorithm 1; b) Gradient Boost; and c) Processing time (per image) on thirty-five test trees across all five weeks of data collection in 2023.}
    \label{fig:validation_graphs}
\end{figure}

\begin{figure}[!h]
    \centering
    \includegraphics[width=15cm]{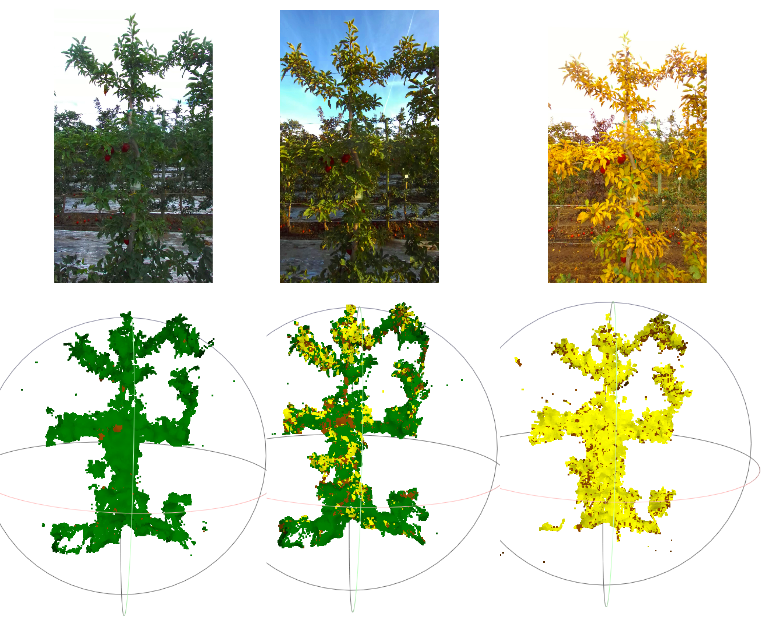}
    \caption{The images (top row) and segmented point cloud (bottom row) of a test tree during weeks 1,3, and 5. The three colors in the point cloud \textcolor{red}{(bottom row)} represent three groups; green-green leaves, yellow-yellow leaves, and brown-trunk, branches and some leaves that have turned brown.}
    \label{fig:weeks-clustering}
\end{figure}

Using the validation data (detailed in Section \ref{sec:clustering}), the Gradient Boost model, with the optimal hyperparameter configuration, was compared against Algorithm 1.  Figure \ref{fig:yellowness_valid} qualitatively shows the performance of the two models within the cropped canopy regions (between the second and fourth trellis wire). Quantitative assessment of the performance of these models was performed using the Yellowness Indices (calculated with the proposed method as well as the manual estimation in the field) of the validation trees (Figure \ref{fig:validation_graphs}). Both models were similar  in their classification accuracy on the validation dataset with $R^2$  of 0.69 for Algorithm 1 and 0.72 for gradient boost model. This suggests  there is a good fit between the ground truth (Yellowness indices of the validation trees) and calculated yellowness. However, Algorithm 1 had significantly slower processing speeds compared to the Gradient Boost method. On the same point cloud, the Gradient Boost model was approximately six times faster (with average segmentation time of 0.39 sec per tree) compared to Algorithm 1 (2.41 sec per tree). Consequently, Gradient Boost was used for subsequent point cloud segmentations. Figure \ref{fig:weeks-clustering} demonstrates the results from the gradient boost segmentation model for a test tree during weeks 1, 3, and 5.

\subsection{\textit{\textit{Yellowness Index}} and Leaf Nitrogen Concentration}\label{sec:nitrogen_realtion_result}

\begin{figure}[!h]
    \centering
    \includegraphics[width=\textwidth]{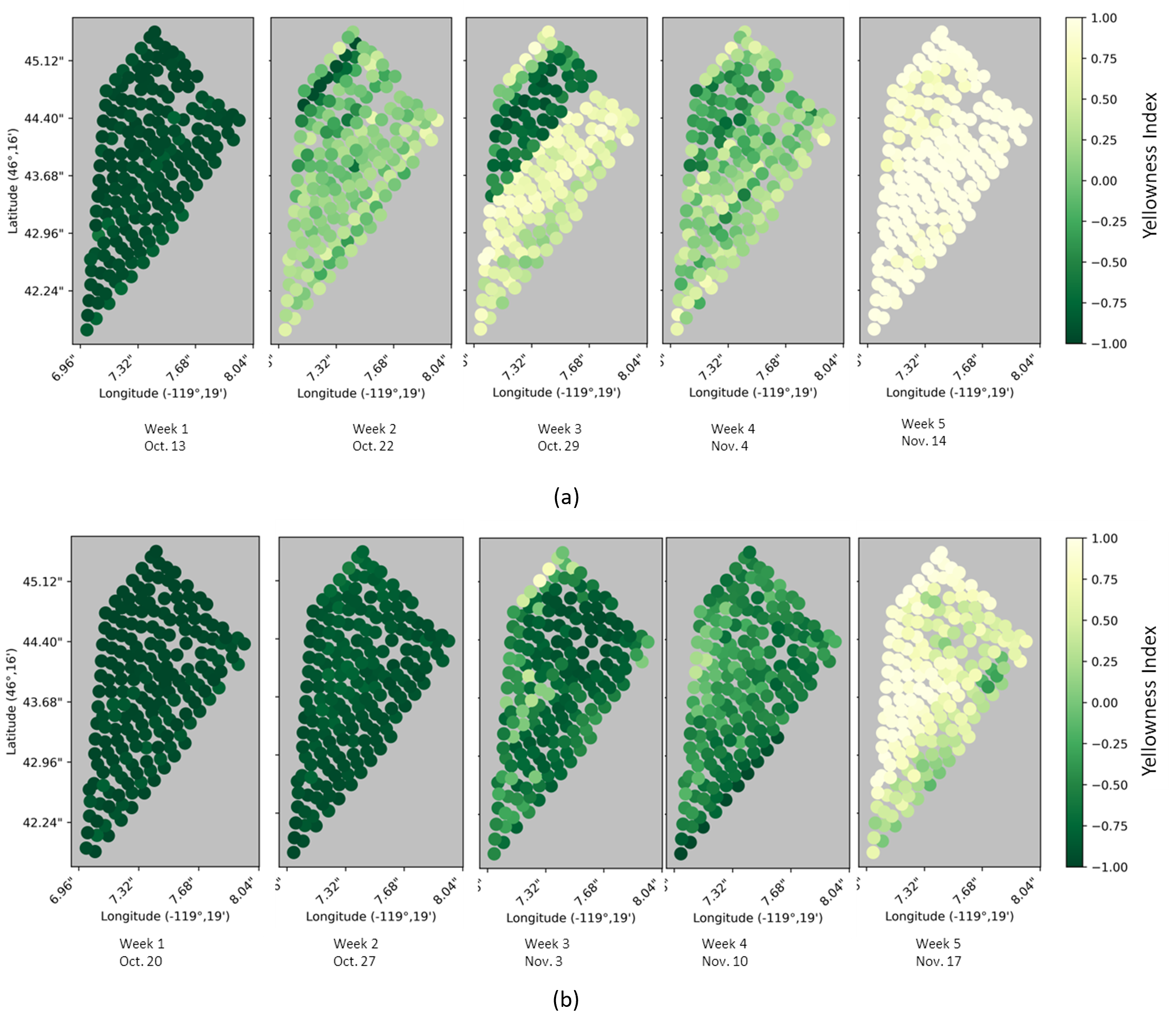}
    \caption{Spatial variation of \textit{yellowness index} over the study period in the test orchard; a) 2021 and b) 2023 datasets. The color bar on the right indicates the yellowness level, with darker shade representing greener trees and lighter shade representing yellower trees..}
    \label{fig:spatial}
\end{figure}

The segmented point clouds were used to calculate the \textit{yellowness index} for each tree over individual weeks using equation \ref{eq:yel}.  The pattern and progression of yellowing of trees  is depicted in spatial maps of these yellowness indices during fall 2021 (Figure \ref{fig:spatial}a) and fall, 2023 (Figure \ref{fig:spatial}b). The darker points in these maps represent greener trees and the lighter points represent trees with more yellow leaves.  In both years we recorded a consistent increase in \textit{yellowness index} over time. However, the progression of yellowing was different between the two years. In 2021, trees began the  transition to yellow in week 2 (around October 22), whereas in 2023, this change occurred later in the year  after November 10. This difference may be due to relatively higher temperature during the growing season in 2021 compared to 2023 (AgWeatherNet). Comparing years in growing degree days (GDD) provides a more precise measure of plant growth stages than calendar days, as GDDs account for the variable influence of temperature on developmental rates, allowing for more accurate predictions of phenological events regardless of annual temperature fluctuations \citep{dass2013growing}. Prior studies by De la \citet{de2014high} and \citet{kim2020high} have demonstrated that elevated temperatures can induce early senescence in foliage.  In our research site, the timing of full bloom  in 2021 was almost three weeks earlier  than 2023.  This difference in bloom correlates with the observed variations in the timing of tree yellowing, and further highlights the role of temperature influencing key phenological stages. In both years, yellowing occurred approximately 29 weeks after full bloom.

The correlation between leaf nitrogen concentration and yellowness index was determined for each week in 2021 and 2023. The trees were categorized into five groups based on their leaf nitrogen concentrations. In general, tree nitrogen levels were poorly correlated with the yellowness index in both years (r values ranging from -0.28 to -0.43 in 2021 and 0.16 to -0.64 in 2023). However, an extended retention of greenness was observed in trees that had high N content compared to those with lower N levels. Previous studies have shown strong correlations between color indices and leaf nitrogen concentration in various plant species. For example, \citet{lee2019field} reported a correlation coefficient of 0.85–0.91 between SPAD readings and leaf nitrogen concentration in apples, while \citet{el2018inspect} found an R² of 0.81 for citrus trees. \citet{haider2021computer} observed an even higher $R^2$ of 0.91 across multiple crops, though their system required crop-specific calibration.

Additional studies using hyperspectral imaging have also demonstrated higher predictive accuracy. \citet{ye2020estimation} reported R² values of 0.77–0.78 in estimating leaf nitrogen concentration using both PLS and MLR models, based on raw reflectance and its first derivative, at both leaf and canopy levels. \citet{jang2023estimation} used hyperspectral imaging combined with SVR, PLSR, and XGBoost, and maintained similar predictive performance even when reducing spectral resolution through binning strategies, with R² values comparable to full-spectrum models. \citet{li2022inversion} achieved a validation R² of 0.75 using UAV-acquired hyperspectral imagery and a backpropagation neural network, indicating the potential of aerial platforms for canopy-level nitrogen mapping. Similarly, \citet{chen2020rapid} reached an R² of 0.843 by combining canopy hyperspectral reflectance with ensemble learning techniques, using extensive preprocessing including SNV-FD and feature selection methods . \citet{wen2018nitrogen} also reported an $R^2$ of 0.72 using support vector machines on SG-smoothed first derivative spectral data of apple leaves. More recently, \citet{chen2024novel} developed a deep learning framework that incorporated phenological data - days after anthosis along with hyperspectral reflectance to estimate leaf nitrogen content, achieving validation R² improvements from 0.69 to 0.79 while eliminating 96\% of redundant spectral bands.

In contrast, the method evaluated in this study achieved an $R^2$ of 0.72 for estimating the yellowness index using a gradient boost model. Although the correlation between the yellowness index and leaf nitrogen was generally lower than in prior studies, moderate correlations (r = 0.4–0.6) were found during certain weeks, suggesting that specific time windows exist where color-based estimation of nitrogen status may be more effective. The reduced performance in this study may be attributed to factors such as variable natural lighting, occlusion within the canopy, and a relatively narrow distribution of nitrogen concentration values —particularly in the lower N range.

Nonetheless, the system developed here captured relative differences and temporal trends in canopy color across large numbers of trees and under field conditions. Unlike many hyperspectral approaches, which require costly instrumentation and intensive spectral calibration, this system provides a low-cost, scalable, and non-destructive alternative for monitoring nitrogen dynamics. While higher $R^2$ values were reported in studies relying on hyperspectral reflectance, the integration of point cloud segmentation and color-based metrics in this study offers unique benefits in field-based decision support contexts. The potential utility of the method can be enhanced by combining the yellowness index with other features such as canopy density, trunk diameter, and phenological data to improve nitrogen estimation in orchards.

\begin{figure}[!h]
    \centering
    \includegraphics[width=0.75\textwidth]{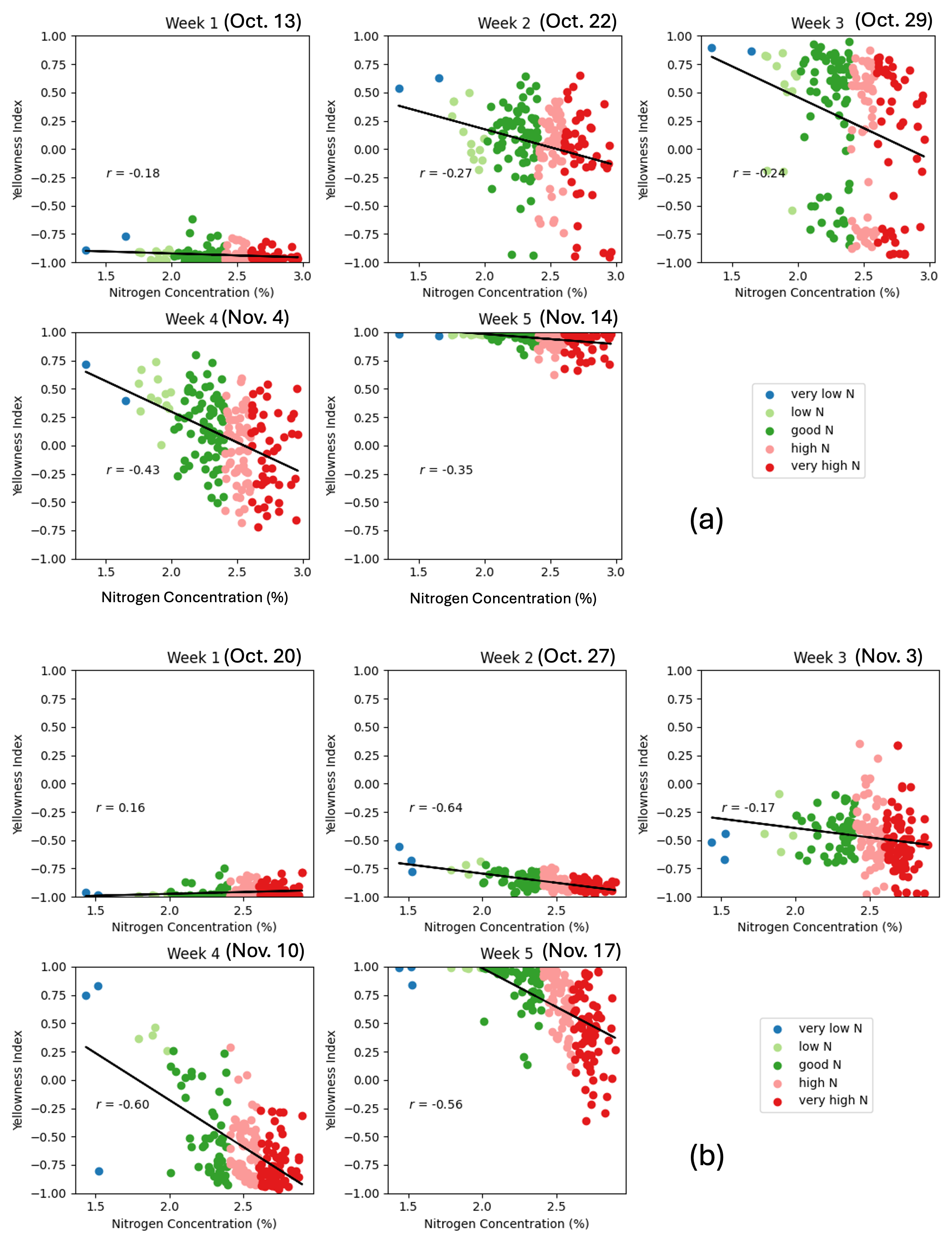}
    \caption{Correlation between N concentration and \textit{\textit{yellowness index}}over weeks during the fall season of (a) 2021 and (b) 2023 in trees shown with Nitrogen \% in the trees. (Very low N ($N<1.7\%$), Low N ($1.7\% < N <2.0\%$), Good N ($2\%< N < 2.4\%$), High N($2.4\%< N <2.6\%$), Very high N ($N>2.6\%)$. The dates on the plots show the date when the data were collected in 2021 and 2023.}
    \label{fig:yellowness}
\end{figure}

\begin{figure}[!h]
    \centering
    \includegraphics[width=0.65\textwidth]{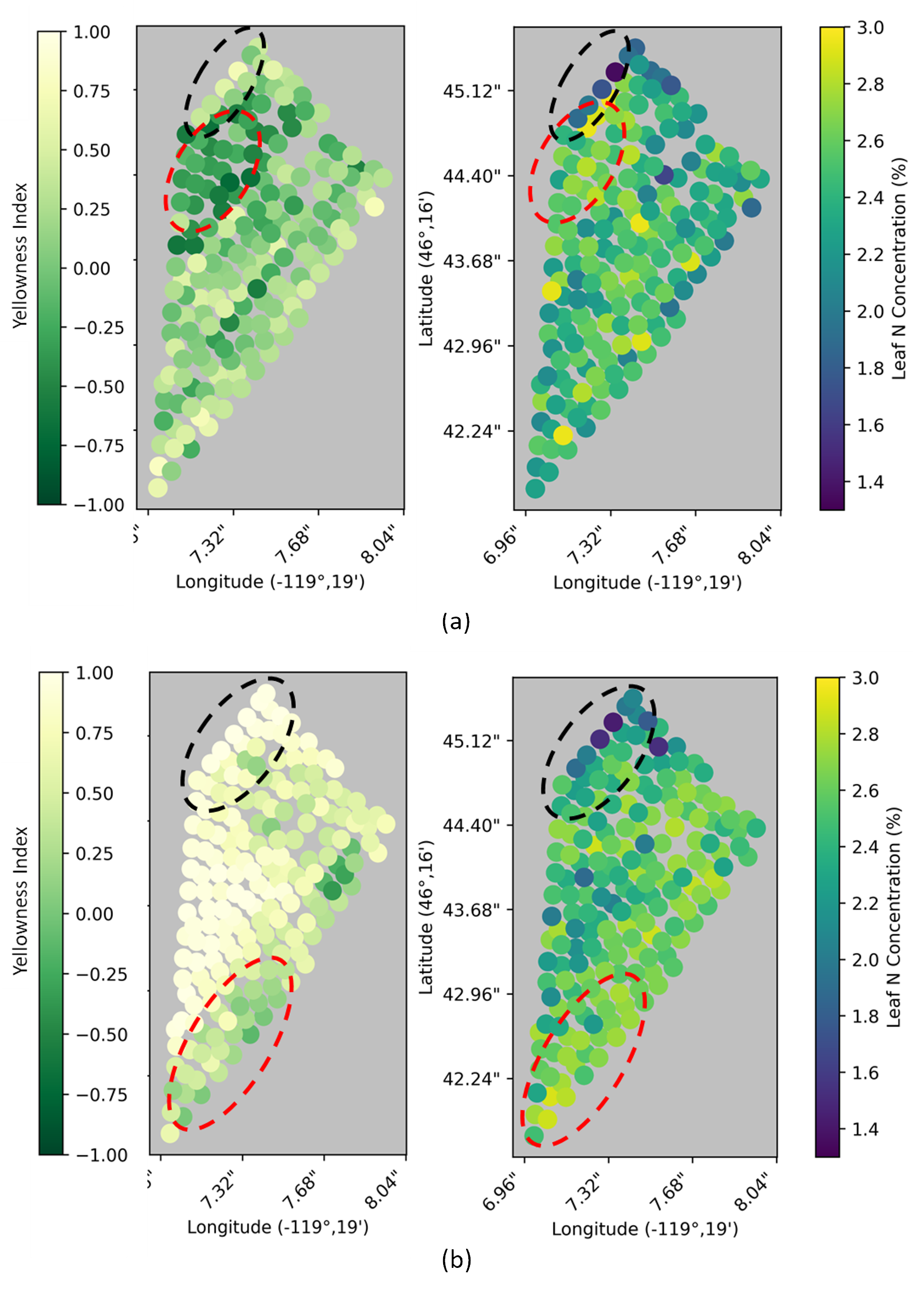}
    \caption{\textit{\textit{yellowness index}} (left plots) variation with leaf nitrogen concentrations (right plots) in (a) week 4 of 2021, and (b)week 5 of 2023. The color bars on the left and right show the values of \textit{\textit{yellowness index}} and leaf nitrogen concentrations. The red ellipses show the sample areas with higher nitrogen concentrations, and the black ellipses show areas with lower nitrogen concentrations. }
    \label{fig:nitrogen_yellowness_comparision}
\end{figure}

The variability among trees in leaf nitrogen concentration and yellowness index is presented in Figure \ref{fig:nitrogen_yellowness_comparision} (a – week 4 of 2021; b – week 5 of 2023). In both plots, red ellipses highlight areas of the field with higher leaf N and black ellipses represent areas with lower leaf N. In both  plots the area with lower leaf Nitrogen  has higher yellowness indices in the left plots and vice versa. 

The results from the one-way ANOVA suggested that a significant difference between the group means existed in weeks 1, 3, and 4 of 2021 and weeks 2, 4, and 5 of 2023, at a significance level of $p < 0.05$. (Table \ref{table:significant}). To identify which specific nitrogen level trees differed, a subsequent Tukey HSD test was performed.  The columns labeled "Group 1" and "Group 2" show the pairs of trees with those nitrogen levels exhibited significant differences during each week. Each row in the table represents a significant difference in group means between the values in the "Group 1" and "Group 2" columns for the specified week. Significant differences in \textit{yellowness index} between trees with different nitrogen levels were most prominent during weeks 2 and 4 in 2023 and week 4 in 2021.

\begin{table}[!h]
\centering
  \caption{Post-hoc analysis (Tukey test) between trees at different N concentrations when divided into 3 and 5 classes based on leaf Nitrogen concentrations at a) 2021 and b) 2023. The given rows showed significant differences between Group 1 and Group 2 in the column during the given week at a significance level of $p<0.05$.}
  \label{table:significant}
  \includegraphics[width=0.65\linewidth]{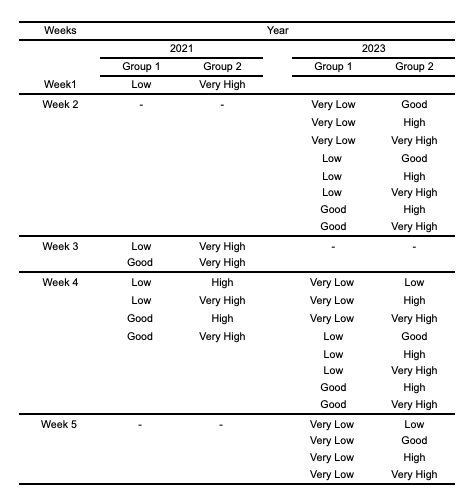}
\end{table}

While this study suggested a generally lower correlation, there were some weeks in which the correlations were moderate (~0.4-0.6), indicating that there may be specific times when trees with different nitrogen concentrations display different \textit{yellowness indices}. Capturing data during these periods could be useful in identifying trees at different nitrogen levels more effectively. To establish the appropriate timing for accurate and robust \textit{yellowness index} assessment, various factors might need to be considered, such as temperature, growing degree days, precipitation, among others.  However, it should be noted that, overall, the correlation remained poor, which suggests that other factors—such as differences in vigor of the trees, irrigation, or the presence of other nutrients—could also have contributed to the variation in \textit{yellowness index}. By identifying these critical time points (in terms of days after full bloom or growing degree days) when significant differences in leaf color occur between trees at different nitrogen concentrations, the yellowness index measured could potentially be used for accurately assessing tree nitrogen status.

The results from this study demonstrated that a machine vision system could be used to identify differences in the yellowing patterns of trees. The outcomes and methodologies of this study offer a means to quantify tree color that could be used to determine tree nitrogen levels, which are crucial factors in guiding fertilizer application to individual trees. However, the system relies on a machine vision system which is susceptible to variations in external lighting conditions. Nevertheless, recent advancements in machine vision sensors have addressed low-light conditions using innovative techniques such as high dynamic range \citep{zhang2020learning} and Adversarial Networks\citep{chen2018deep}. In addition, deep learning methods have notably enhanced imaging system resilience to diverse lighting environments\citep{li2021low}. The algorithm employed in this study relies on labeled data for initially training the model to differentiate yellow and green points on trees, necessitating manual inputs. It’s important to note that this algorithm might require further calibration between different years and across tree varieties for optimal performance. 

\section{Conclusion and Future Work} \label{conclusions}

Leaf Nitrogen concentration is one of the critical factors that determines the yield and quality of fruits and the overall health of apple trees. Growers often rely on different visual cues including the canopy color for assessing plant health and nitrogen needs. This study presented a machine vision-based approach for quantitatively assessing tree canopy color and utilizing the color during transitional period in the fall (when the leaves change color from green to yellow during senescence) to assess N status in trees. Specifically, this study segmented the test tree canopies in an outdoor environment with natural background using 3D point clouds, identified yellow and green regions in the canopy using a gradient boost model, and used a metric called \textit{yellowness index} to quantify the canopy color. It was found that the trees notably began transitioning in color around the $29^th$ week following full bloom in both years the test was conducted. The trees with lower and higher nitrogen showed significant differences between the \textit{yellowness indices} during week 4 (Around November 4) in 2021 and week 2 (October 27) and week 4 (November 10) in 2023. The difference in these dates signifies the importance of factors such as temperature, humidity, and growing degree days during different years in predicting optimal window of estimate \textit{yellowness index} for tree nitrogen levels assessment. Overall, the findings from this study can be summarized as:

\begin{enumerate}
    \item A machine vision-based system can be used to segment yellow and green foliage in apple trees in outdoor orchard environments. A metric \textit{yellowness index} was defined to quantify the status of tree foliage color, which was estimated with a $R^2$ value of 0.72 using a gradient boost classifier model.

    \item The \textit{yellowness index} of the trees was found to be correlated with the leaf nitrogen levels in the trees during some periods of the study. Trees with different nitrogen levels showed different yellowing patterns. The difference between low and high nitrogen trees were more significant ($p<0.05$) in week 4 in 2021(31 weeks after full bloom) and 2023 (29 weeks after full bloom).
\end{enumerate}

This study presented a new approach for assessing the color of apple tree canopies and explored its relationship with leaf Nitrogen concentration. This method could be a good alternative to traditional methods like chemical methods, chlorophyll meter, and spectral analysis of N assessment in apple trees and can give instantaneous results that could be expanded to individual tree level assessment using a ground vehicle or robot. The study also provided critical insights into the fall color changes in apple trees and their relationship with leaf nitrogen concentrations.

One major limitation of assessing leaf nitrogen level using the proposed method is that some years, the tree might freeze early due to rapid decline in temperature and never show the color change. In such cases, it is important to not rely on just a single factor and have a decision support system that takes multiple features as input, making the decision more robust and reliable. The \textit{yellowness index}values along with other commonly used visual features like canopy density, trunk cross-sectional area, and shoot length can be combined into a single robust model and could aid in developing a robust decision support system for efficient fertilization plans tailored to individual tree nitrogen needs.

\section*{Acknowledgement}
This research was supported by the Washington Tree Fruit Research Commission. Special thanks to David Allan for providing access to the test orchard and for valuable feedback and support during this work.


\bibliographystyle{elsarticle-harv}
\bibliography{ref}
\end{document}